\documentclass[runningheads]{llncs}
\usepackage{graphicx}
\usepackage{float}
\usepackage{tikz}
\usepackage{comment}
\usepackage{amsmath,amssymb} 
\usepackage{mathrsfs}
\usepackage{hyperref}
\usepackage{color}
\usepackage{wrapfig}
\usepackage{graphics} 
\usepackage{epsfig} 
\usepackage{times} 
\usepackage{amsmath} 
\usepackage{amssymb}  
\usepackage{subfigure}
\usepackage{booktabs}
\usepackage{multirow}
\usepackage{xcolor}

\usepackage[accsupp]{axessibility}  

\begin{document}
\pagestyle{headings}
\mainmatter
\def\ECCVSubNumber{5923}  

\title{What Matters for 3D Scene Flow Network} 


\titlerunning{What Matters for 3D Scene Flow Network}
%
\author{{Guangming Wang\inst{1} \and
Yunzhe Hu\inst{1} \and
Zhe Liu\inst{2} \and
Yiyang Zhou\inst{3} \and
Masayoshi Tomizuka\inst{3} \and Wei Zhan\inst{3}
 \and
Hesheng Wang\inst{1}}\thanks{Corresponding Author. The first two authors contributed equally.}}
\authorrunning{G. Wang et al.}
%
\institute{Department of Automation, Key Laboratory of System Control and Information Processing of Ministry of Education, Key Laboratory of Marine Intelligent Equipment and System of Ministry of Education, Shanghai Jiao Tong University \and
MoE Key Lab of Artificial Intelligence, Shanghai Jiao Tong University \and
Mechanical Systems Control Laboratory, University of California, Berkeley
\\
\email{\{wangguangming,huyz7830,liuzhesjtu,wanghesheng\}@sjtu.edu.cn}
\\
\email{\{yiyang.zhou,tomizuka,wzhan\}@berkeley.edu}
}
\maketitle

\begin{abstract}
3D scene flow estimation from point clouds is a low-level 3D motion perception task in computer vision. 
Flow embedding is a commonly used technique in scene flow estimation, and it encodes the point motion between two consecutive frames. Thus, it is critical for the flow embeddings to capture the correct overall direction of the motion.
However, previous works only search locally to determine a soft correspondence, ignoring the distant points that turn out to be the actual matching ones. In addition, the estimated correspondence is usually from the forward direction of the adjacent point clouds, and may not be consistent with the estimated correspondence acquired from the backward direction. To tackle these problems, we propose a novel all-to-all flow embedding layer with backward reliability validation during the initial scene flow estimation. Besides, we investigate and compare several design choices in key components of the 3D scene flow network, including the point similarity calculation, input elements of predictor, and predictor \& refinement level design. After carefully choosing the most effective designs, we are able to present a model that achieves the state-of-the-art performance on FlyingThings3D and KITTI Scene Flow datasets. Our proposed model surpasses all existing methods by at least 38.2\% on FlyingThings3D dataset and 24.7\% on KITTI Scene Flow dataset for EPE3D metric. We release our codes at \href{https://github.com/IRMVLab/3DFlow}{https://github.com/IRMVLab/3DFlow}.
\end{abstract}
\section{Introduction}
As a fundamental task in computer vision, scene flow estimation aims to estimate a 3D motion field consisting of point-wise or pixel-wise 3D displacement vectors between two consecutive frames of point clouds or images. It provides a low-level representation and understanding of the motion of dynamic objects in the scene. Many applications directly benefit from the techniques used in scene flow estimation, such as semantic segmentation \cite{liu2019meteornet}, {multi-object tracking \cite{wang2020pointtracknet,wang2022interactive}, point cloud registration \cite{liu2019flownet3d,wang2021pwclo,wang2021efficient}}, etc. The performance of scene flow estimation algorithms on point clouds has been greatly improved since deep learning was first applied in \cite{liu2019flownet3d}. Recent studies \cite{liu2019flownet3d,gu2019hplflownet,wu2020pointpwc,puy2020flot,kittenplon2021flowstep3d,li2021hcrf,wang2021hierarchical,wang2022sfgan} focus more on estimating scene flow in an end-to-end fashion from two consecutive frames of raw 3D point clouds. These approaches predict scene flow with only 3D coordinates of point clouds as inputs with no need for any prior knowledge of the scene structure. This paper also focuses on such a research topic.

Previous learning-based methods \cite{liu2019flownet3d,wu2020pointpwc,wang2021hierarchical} adopt flow embedding to correlate adjacent frames of point clouds and to encode point motion. Their models then propagate the flow embedding through set upconv layers \cite{liu2019flownet3d} or coarse-to-fine warping \cite{wu2020pointpwc,wang2021hierarchical} to regress the scene flow. FlowNet3D \cite{liu2019flownet3d}, for example, introduces the flow embedding in a point-to-patch manner, which means that a specific point in the first point cloud $PC_1$ merely uses several neighbouring points in the second point cloud $PC_2$ to learn the correlation. PointPWC-Net \cite{wu2020pointpwc} further improves it and proposes to learn a patch-to-patch flow embedding, which adds a second point-to-patch embedding process in $PC_1$ itself after the first point-to-patch embedding between two point clouds. In addition, Pyramid, Warping,
and Cost volume (PWC) structure \cite{sun2018pwc} is introduced to refine the scene flow for several times.
HALFlow \cite{wang2021hierarchical} also follows this PWC structure \cite{sun2018pwc} but improves by introducing the attention mechanism in both embedding processes. 
However, during the first embedding process between two point clouds, these methods only search for $K$ Nearest Neighbours (KNN) in $PC_2$ to aggregate correspondence information. Practically, $K$ is substantially smaller than the total number of points in the second frame, making it possible for a point in $PC_1$ to miss the correct yet distant matching point in $PC_2$. 
Moreover, it is extremely important to obtain a reliable correlation when calculating it for the first time because it encodes the overall direction of the flow. The scene flow will be eventually misguided if this issue is ignored.
To tackle this problem, we introduce a novel all-to-all flow embedding layer based on the double attentive flow embedding layer in HALFlow \cite{wang2021hierarchical}. With all-to-all embedding, each point in $PC_1$ will use all points in $PC_2$ for correlation during the first embedding process, and each point in $PC_2$ can therefore obtain the correlation with all points in $PC_1$ too. This mechanism allows that the feature correlation of all points can be exhaustively utilized from both sides and reliable correspondence estimation can be further achieved.

This all-to-all mechanism, however, cannot guarantee that the reliable correlation is bi-directional. That is, the estimated match pair for a specific point in $PC_1$ in the forward direction may not be consistent with the match for the corresponding matched point in $PC_2$ in the backward direction. Therefore, we need another constraint on the backward match to validate its consistency with the forward pass. Mittal \emph{et al}. \cite{mittal2020just} utilize a similar mechanism by designing a cycle-consistency loss to achieve self-supervising the scene flow estimation. However, we expect to directly incorporate this constraint of backward validation into our network to allow the network to learn this ability in forward reasoning. To this end, we propose backward reliability validation, a joint learning method of backward constraint in the all-to-all flow embedding layer.

Furthermore, there are several components of our network with alternative designs either from themselves or from other works that could affect the performance. Therefore, we conduct a series of ablation studies to compare different designs and to explore which elements are important and which designs are suitable for 3D scene flow network, including the point similarity calculation, predictor elements choice, and predictor \& refinement level design. Our key contributions are as follows:
\begin{enumerate}
    \item A novel all-to-all point mixture module with backward reliability validation is proposed for reliable correlation between point clouds. The all-to-all mechanism is adopted to capture reliable match candidates from the distance, and backward information is integrated in the inference process to validate the matching consistency.
    
    \item Different designs and techniques of 3D scene flow network are widely compared and analyzed. \emph{Point Similarity Calculation}, \emph{Designs of Scene Flow Predictor}, \emph{Input elements of Scene Flow Predictor}, and \emph{Flow Refinement Level Design} are individually discussed and evaluated to showcase what matters in 3D scene flow network.
    
    \item Experiments demonstrate that our model achieves state-of-the-art performance, reducing EPE3D metric by at least 38.2\% on FlyingThings3D dataset \cite{mayer2016large} and 24.7\% on KITTI Scene Flow dataset \cite{menze2018object}. The effectiveness of proposed techniques and choices of network designs are demonstrated through extensive ablation studies.   
\end{enumerate}
\section{Related Work}

The concept of scene flow is first introduced by Vedula \emph{et al}. \cite{vedula1999three} as the 3D motion field in real-world scenarios. Many previous works estimate scene flow by recovering the 3D motion from optical flow and depth information on 2D image pairs, either using RGB-stereo \cite{huguet2007variational,pons2007multi,valgaerts2010joint,menze2015object,vogel2013piecewise,vogel20153d,mayer2016large,ma2019deep} or RGB-D \cite{hadfield2011kinecting,herbst2013rgb,jaimez2015primal} data. {There has also been some recent works focusing on recovering scene flow from monocular camera \cite{yang2020upgrading,hur2020self,yang2021learning,hur2021self,wang2021unsupervised}.} However, since scene flow indicates the 3D motion, directly estimating scene flow from 3D data input can enable direct optimization and higher accuracy. The applications of LiDARs in recent years have created more available raw data of point cloud, and point-cloud-based scene flow estimation approaches \cite{dewan2016rigid,ushani2017learning} are rapidly emerging. 

{Since deep learning has shown excellent performance for raw point-cloud-based tasks \cite{wang2021anchor,wang2021spherical,liu2019lpd,xia2021soe} with the proposal of PointNet \cite{qi2017pointnet} and PointNet++ \cite{qi2017pointnet++},} many works estimate scene flow directly from raw point clouds in an end-to-end fashion. FlowNet3D \cite{liu2019flownet3d} presents the first end-to-end scene flow estimation framework on point clouds. It uses PointNet++ \cite{qi2017pointnet++} to extract local point features and introduces a flow embedding layer to encode the point motions. 
HPLFlow-Net \cite{gu2019hplflownet} leverages the idea from Bilateral Convolutional Layers (BCL) \cite{kiefel2015permutohedral,jampani2016learning} and proposes DownBCL, UpBCL, and CorrBCL designs to restore structural information of large-scale point clouds. 

More recent works focus on improving the network performance through introducing new techniques or incorporating new components. FLOT \cite{puy2020flot} proposes to find the correspondences from an optimal transport module by graph matching. 
PointPWC-Net \cite{wu2020pointpwc} follows a coarse-to-fine fashion for scene flow estimation on point clouds. It extends the important component of cost volume in optical flow network \cite{sun2018pwc} and proposes a novel point-based patch-to-patch cost volume. HALFlow \cite{wang2021hierarchical} improves the aforementioned cost volume by a novel double attentive flow embedding method that distributes more weights on task-related regions. HCRF-Flow \cite{li2021hcrf} focuses on maintaining the local geometric smoothness with the help of Conditional Random Fields (CRFs) in deep neural networks and proposes a high-order CRFs module as the formulations of spatial smoothness and rigid motion constraints. FESTA \cite{wang2021festa} improves naive Farthest Point Sampling (FPS) by proposing a trainable Aggregate Pooling (AP) to adaptively shift points to invariant positions.  Inspired by \cite{puy2020flot}, FlowStep3D \cite{kittenplon2021flowstep3d} designs a Global Correlation Unit that computes a soft correlation matrix to guide the initially estimated scene flow and adopts Gated Recurrent Unit (GRU) for local flow update. PV-RAFT \cite{wei2021pv} leverages point-based and voxel-based features and presents point-voxel correlation fields to capture both local and long-range dependencies for point pairs. 

In \cite{wu2020pointpwc,kittenplon2021flowstep3d}, different information is utilized for updating local scene flow but they do not show which information is more important for the input of the updating unit. In \cite{kittenplon2021flowstep3d,wei2021pv}, GRU is used for iterative flow update inspired by RAFT \cite{teed2020raft} and claimed to be more effective than a fully-connected layer. Since RAFT \cite{teed2020raft} has shown a promising performance on optical flow, we also want to know whether the use of GRU will improve the performance in other scene flow network structures. In \cite{puy2020flot,kittenplon2021flowstep3d}, element-wise product and cosine similarity are used to represent correlation between points while concatenation of point feature is implemented for learning correlation in \cite{wang2021hierarchical}, but none of them gives evaluation about which one is better. This paper will discuss the above issues and compare what matters for 3D scene flow network based on PWC structure.
\section{3D Scene Flow Network}

\begin{figure*}[t]
	\centering
	\includegraphics[width=1.00\linewidth]{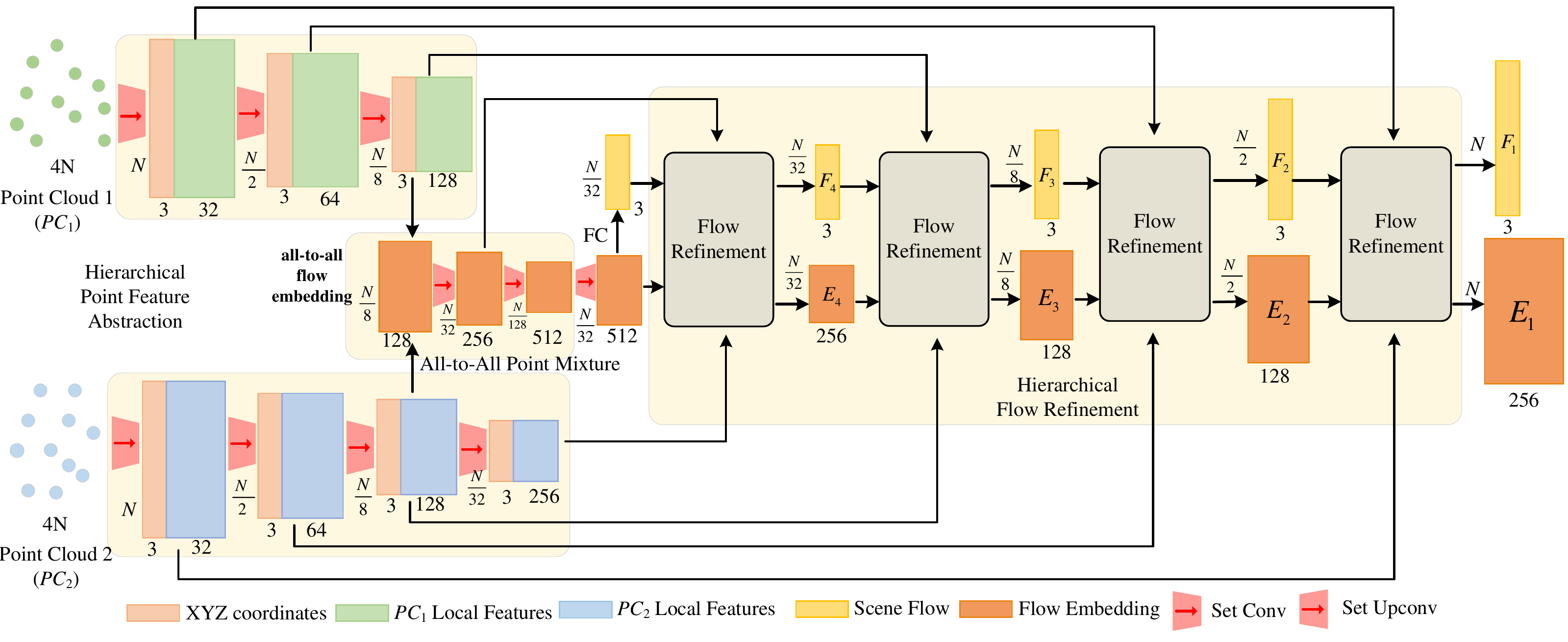}
	\caption{The detailed architecture of our network. Three set conv layers for $PC_1$ and four set conv layers for $PC_2$ constitute the hierarchical point feature abstraction module. The all-to-all point mixture module consists of one layer of all-to-all flow embedding followed by two set conv layers. Four flow refinement layers are constructed in the hierarchical flow refinement module.}   
	\label{figure:network}
\end{figure*}
\subsection{Network Architecture}
Our proposed network, illustrated in Fig. \ref{figure:network}, takes in two frames of point clouds with 4$N$ points in each, which are $PC_1$ and $PC_2$, and estimates $N$ points' scene flow from coarse to fine. Our network is comprised of three modules: 1) Hierarchical Point Feature Abstraction, 2) All-to-All Point Mixture, and 3) Hierarchical Flow Refinement. 

The hierarchical point feature abstraction module has three set conv layers from \cite{liu2019flownet3d} for $PC_1$ and four set conv layers for $PC_2$. Each set conv layer performs down-sampling operation on the input points and extracts local features of the down-sampled points. The same level of the set conv layers shares the same weights. Then, the proposed all-to-all flow embedding layer correlates two point clouds and learns the flow embedding. 
We then use two set conv layers after the flow embedding layer for smoothness. Next, the output of the all-to-all point mixture module is up-sampled by the set upconv layer from \cite{liu2019flownet3d} to generate the initial flow embedding. A Fully-Connected (FC) layer is thereafter applied on the initial flow embedding to produce the initial scene flow. Finally, the initial scene flow and flow embedding are both fed into the hierarchical flow refinement module and refined iteratively to derive the final scene flow using the information from specific level. The skip connections indicate which level of information is utilized. 

\subsection{Hierarchical Point Feature Abstraction}
\label{section:Hierarchical Point Feature Abstraction}
In the hierarchical point feature abstraction module, two consecutive point clouds are down-sampled and encoded through a series of set conv layers respectively. {We adopt the set conv layer in PointNet++ \cite{qi2017pointnet++} to perform point feature abstraction}.

Each set conv layer consumes $n$ points $\{(x_i, p_i)~|~i=1,\dots,n\}$, {where $x_i \in \mathbb{R}^3$ and $p_i \in \mathbb{R}^c$ represent 3D coordinate and the point feature}. The output of each layer are $n'$ sampled point $\{(x_j',p_j')~|~ j=1,\dots,n' \}$ with {$x_j' \in \mathbb{R}^3$ and $p_j' \in \mathbb{R}^{c'}$ denoting the 3D coordinate and extracted local feature}. All of the output $n' ~(n'<n)$ are sampled from the input $n$ points using Farthest Point Sampling (FPS) \cite{qi2017pointnet++}. 

For each of the $n'$ sampled point, its $K$ nearest neighbours $\{(x_i^k, p_i^k)~|~k=1,\dots,K \}$ are selected from the input $n$ points. Then, a learnable shared Multi-Layer Perceptron (MLP) and max-pooling operation are adopted to extract the point feature $p_j'$ of each sampled point from the $K$ neighbouring points. The point feature $p_j'$ is formulated as:
\begin{equation}
    \label{eq1}
    p_j' = \mathop{\text{maxpool}}_{k=1,\dots,K}(\text{MLP}((x_j^k-x_j')\oplus p_j^k)),
\end{equation}
where $\oplus$ indicates concatenation operation. $\mathop{maxpool}$ means max-pooling operation.
\subsection{All-to-All Point Mixture}
\label{section:all2all}
The inputs of the all-to-all embedding layer are two consecutive frames of point clouds: $PC_1 = \{(x_i,p_i)~|~i=1,\dots,n_1\}$ and $PC_2 = \{(y_j,q_j)~|~j=1,\dots,n_2\}$, sampled in the hierarchical point feature abstraction module. $x_i,y_j \in \mathbb{R}^3$ indicate 3D coordinates and $p_i,q_j \in \mathbb{R}^c $ indicate the point feature. The output of the layer will be the flow embedding $E = \{e_i ~|~ e_i \in \mathbb{R}^c , i=1\dots,n_1\}$, which utilizes exhaustive information in two point clouds and encodes motion for points in $PC_1$.

\begin{figure*}[t]
	\centering
	\includegraphics[width=1.00\linewidth]{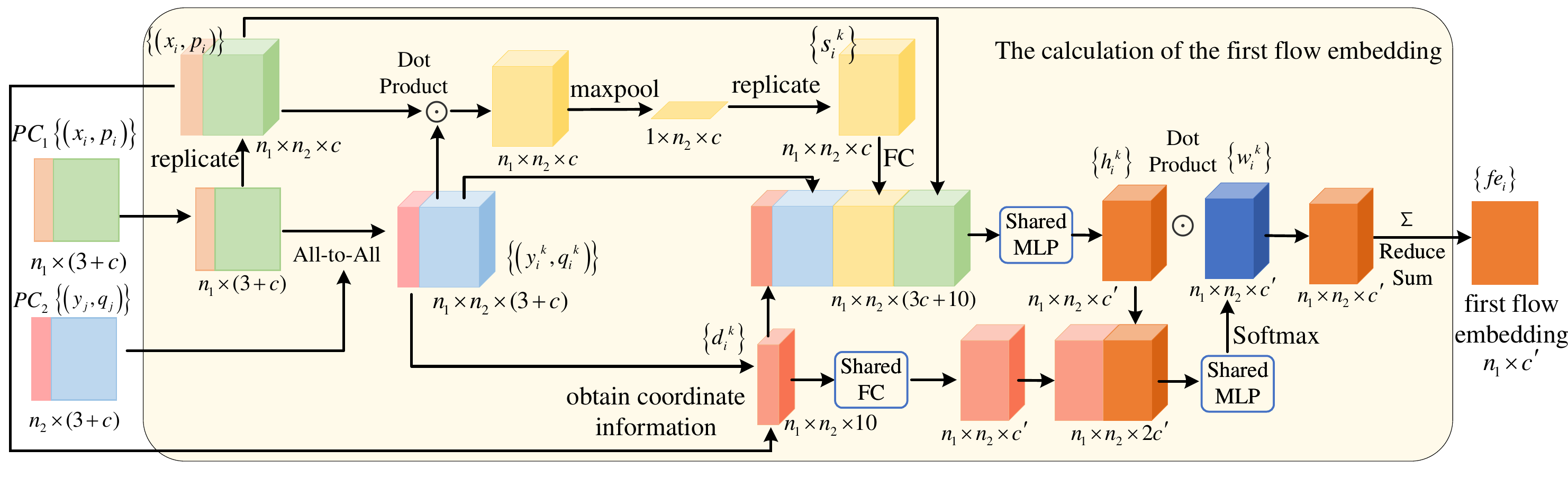}
	\caption{The detailed calculation of the first flow embedding between $PC_1$ and $PC_2$.}   
	\label{figure:all2all cost volume}
\end{figure*}

Our all-to-all embedding includes a two-stage attention-based embedding process with an improved first embedding stage. For the first embedding process, we correlate points in two point clouds by incorporating the all-to-all mechanism with backward reliability validation as shown in Fig. \ref{figure:all2all cost volume}. 
Instead of choosing only $K (K<n_2)$ nearest points, each point in $PC_1$ selects all $n_2$ points {$Q_i = \{(y_i^{k},q_i^{k})~|~ k=1,\dots,n_2\}$} from $PC_2$. {In this process}, all $n_2$ points {$Q_i$} from $PC_2$ are utilized to embed point motion into points in $PC_1$. The motion embeddings will then be updated by carefully designed attentive weighting to derive the first flow embedding {$FE =\{fe_i ~|~ i=1,\dots,n_1 \}$}. The calculation details are elaborated as below.

A 10-dimensional vector capturing the 3D Euclidean space information is first calculated as follows:
\begin{equation}
    {d_i^{k} = x_i \oplus y_i^{k} \oplus (x_i-y_i^{k}) \oplus \left\| x_i-y_i^{k} \right \|_2},
\end{equation}
where $\left\| \cdot \right \|_2$ indicates the $L_2$ norm. Then, to realize backward reliability validation, we first formulate a vector that represents a form of similarity between two point clouds by applying element-wise product of $PC_1$ point feature {$p_i$} and $PC_2$ point feature {$q_i^{k}$}. Max-pooling operation is then performed over $n_1$ candidate backward embedding features, {selecting the most potential and reliable matching candidates in $PC_1$ for each point of $PC_2$}. An FC layer is then adopted to encode the backward reliability information. The calculation of backward validation vector is as follows:
\begin{equation}
    {s_i^{k} = \text{FC}(\mathop{\text{maxpool}}_{i=1,\dots,n_1}(p_i \odot q_i^{k}))},
\end{equation}
where $\odot$ denotes dot product. The first flow embedding before attentive weighting is then formulated as:
 \begin{equation}
    \label{eq:similarity}
    {h_i^{k} = \text{MLP}(d_i^{k} \oplus p_i \oplus q_i^{k} \oplus s_i^{k})}.
\end{equation}

Specifically, {$p_i$} and {$q_i^{k}$} are normalized on the feature channel before concatenation. Given the 10-dimensional vector {$d_i^{k}$}, the first attentive weights for soft aggregation of the queried points can be written as:
\begin{equation}
    {w_i^{k} = \mathop{\text{softmax}}_{k=1,\dots,n_2}(\text{MLP}(\text{FC}(d_i^{k}) \oplus h_i^{k}))}.
\end{equation}

The first flow embedding {$FE =\{fe_i ~|~ i=1,\dots,n_1 \}$} corresponding to points with {$x_i$} coordinates is calculated as:
\begin{equation}
    {fe_i = \sum_{k=1}^{n_2} h_i^{k} \odot w_i^{k}}.
\end{equation}

For the second flow embedding process, we follow the same process as \cite{wang2021hierarchical}, which is an aggregation process within the $PC_1$ self with attention. Each point in $PC_1$ selects several nearest neighbours in $PC_1$ self, and the neighbourhood flow embeddings in {$FE$} will be aggregated into {each} point in $PC_1$ to obtain the second flow embedding $E = \{e_i ~|~i=1,\dots,n_1 \}$, {which is the output of the all-to-all flow embedding layer}.

\emph{Calculation of Point Similarity}
In formula (\ref{eq:similarity}), point feature {$p_i$} of $PC_1$ and {$q_i^{k}$} of $PC_2$ are concatenated to learn the similarity between points from two point clouds. However, there are other ways \cite{puy2020flot,kittenplon2021flowstep3d,jonschkowski2020matters} to calculate and represent the point similarity: 1) product similarity: 
the direct dot product of {$p_i$} and {$q_i^{k}$} as ${sim(p_i,q_i^{k}) = <p_i,q_i^{k}>}$, 2) cosine product similarity: the dot product of {$p_i$} and {$q_i^{k}$} divided by their respective $L_2$ norm as ${sim(p_i,q_i^{k}) = <\frac{p_i}{\|p_i\|_2},\frac{q_i^{k}}{\|q_i^{k}\|_2}>}$, and 3) normalized product similarity: the dot product of {$p_i$} and {$q_i^{k}$} normalized by their respective mean value $\mu$ and standard deviation $\sigma$ over each feature dimension as ${sim(p_i,q_i^{k}) = <\frac{p_i-\mu_i}{\sigma_i},\frac{q_i^{k}-\mu_i^{k}}{\sigma_i^{k}}>}$. We intend to explore whether the concatenation of point feature is more suitable to represent similarity in our network compared with the product similarity presented above. By replacing the concatenation of feature with different forms of product of feature in formula (\ref{eq:similarity}), the effectiveness of our design is demonstrated in experiments.

\subsection{Hierarchical Flow Refinement}
\begin{wrapfigure}{!tr}{0.5\columnwidth}
	\centering
	\includegraphics[width=1.00\linewidth]{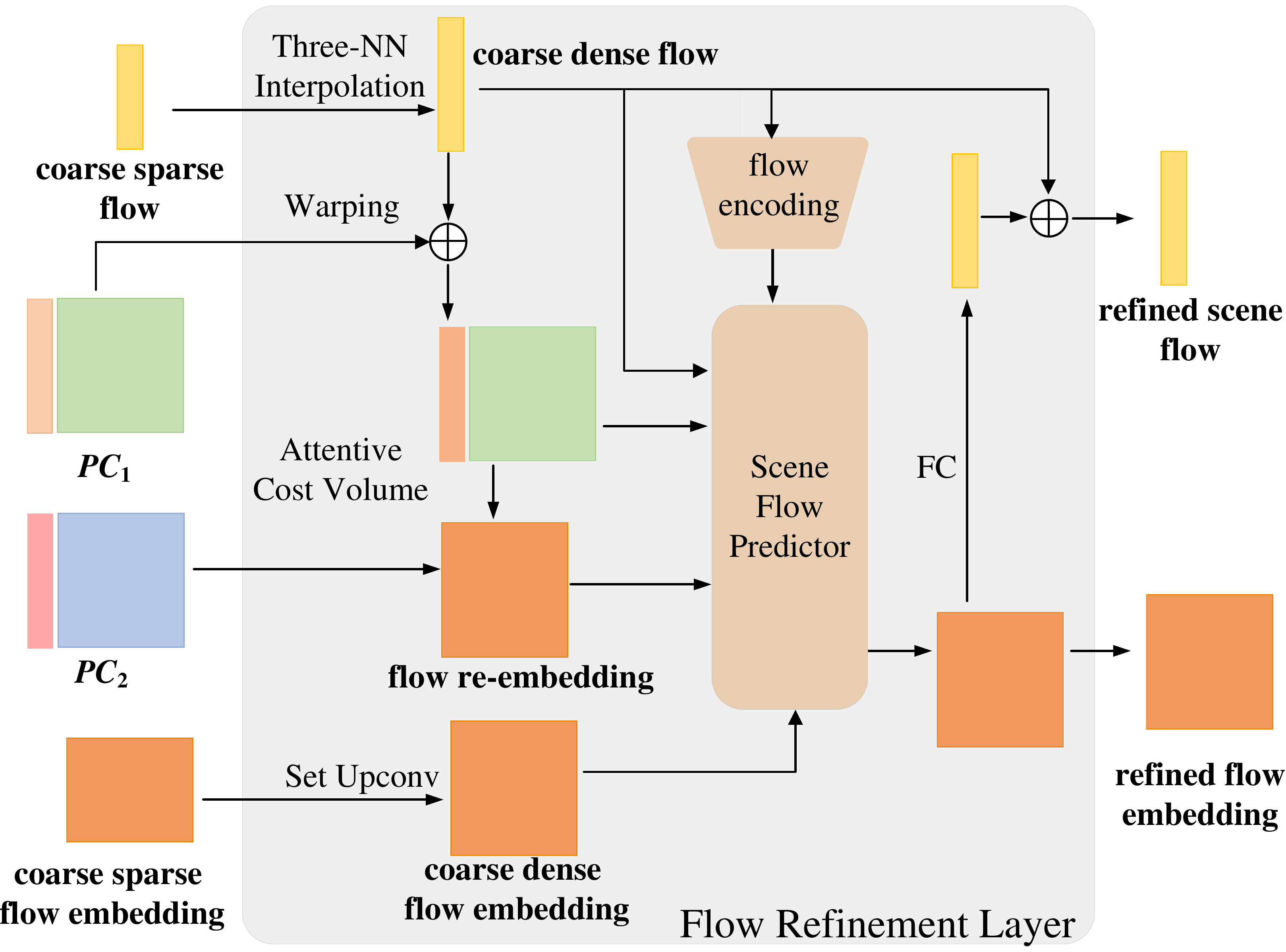}
	\caption{The details of flow refinement layer.} 
	\label{figure:flow-refinement}
\end{wrapfigure}

The hierarchical flow refinement module consists of four flow refinement layers. The layer takes coarse sparse flow and coarse sparse flow embedding as inputs with information of $PC_1$ and $PC_2$ from the previous level while producing refined flow and refined flow embedding as outputs, as illustrated in Fig. \ref{figure:flow-refinement}. It contains four main components: 1) Set Upconv Layer, 2) Position Warping Layer, 3) Attentive Flow Re-embedding Layer, and 4) Scene Flow Predictor. For the first flow refinement layer, the set upconv layer is eliminated to keep the point number unchanged for suitable multi-scale supervision.

\subsubsection{Set Upconv Layer}
\label{section:Set Upconv Layer}
In order to up-sample the coarse sparse flow embedding, the set upconv layer in \cite{liu2019flownet3d} is adopted here to propagate flow embedding from sparse level to dense level. The inputs of this layer are $n$ points with coarse sparse flow embedding $\{(x_i,se_i)~|~se_i \in \mathbb{R}^{d_{sparse}},i=1,\dots,n\}$ and $n' ~(n'>n)$ points with feature $\{(x_j',p_j')~|~j=1,\dots,n' \}$ from the previous dense level. The outputs are $n'$ points with dense flow embedding $\{(x_j',de_j)~|~de_j \in \mathbb{R}^{d_{dense}},j=1,\dots,n' \}$. Specifically, each of the $n'$ dense points will select its KNN from the sparse $n$ points, and the coarse sparse flow embedding will be aggregated to learn the coarse dense flow embedding by MLP.
\subsubsection{Position Warping Layer}

As a coarse-to-fine style, coarse dense flow $\{f_i^{dense}$ $~|~ i=1,\dots,n_1\}$ is first obtained from coarse sparse flow through Three-Nearest Neighbours (Three-NN) interpolation. Next, the coordinates of the first point cloud $PC_1=\{(x_i,p_i)$  $~|~i=,\dots,n_1 \}$ are updated by warping $PC_1$ with coarse dense flow. The warped $PC_1$ is signified as $PC_1'=\{(x_i',p_i)~|~i=1,\dots,n_1 \}$, where $x_i'=x_i+f_i^{dense}$.
\subsubsection{Attentive Flow Re-embedding Layer}
\textcolor[RGB]{0,0,0}{The attentive flow embedding layer proposed in} \cite{wang2021hierarchical} is applied to derive a new flow re-embedding $\{re_i ~|~ re_i \in \mathbb{R}^{d_{re}}, i=1,\dots,=n_1\}$ between $PC_1'$ and $PC_2$. Here, the flow re-embedding contains the flow encoding from each point in $PC_1'$ toward $PC_2$, which is essential in the subsequent refinement.
\subsubsection{Scene Flow Predictor}
\textcolor[RGB]{0,0,0}{The scene flow predictor aims to refine coarse dense flow embedding for the input of later flow refinement layer}. It takes five elements as inputs: 1) the up-sampled coarse dense flow embedding $de_i \in \mathbb{R}^{d_{dense}}$, 2) the flow re-embedding $re_i \in \mathbb{R}^{d_{re}}$, 3) the point feature of the first point cloud $p_i \in \mathbb{R}^{d_{pc_1}}$, 4) the coarse dense flow $f_i^{dense} \in \mathbb{R}^3$, and 5) the dense flow feature $f_i^{enc.} \in \mathbb{R}^{d_{enc.}}$. Specifically, the coarse dense flow is encoded by two set conv layers \cite{liu2019flownet3d} to derive the dense flow feature $f_i^{enc.}$, but {the number of points remain unchanged} instead of being down-sampled. The refined dense flow embedding is formulated as:
\begin{equation}
    \label{eq2}
    de_i' = \text{MLP}(de_i \oplus re_i \oplus p_i \oplus f_i \oplus f_i^{enc.}).
\end{equation}

Finally, we adopt a residual flow learning structure to estimate the refined scene flow. To be specific, an FC layer is first applied on the refined flow embedding to produce the residual flow $f_i^{res}$. Next, the refined scene flow is generated by adding $f_i^{res}$ to $f_i^{dense}$. The calculation of refined scene flow is as follows:
\begin{align}
    \label{eq3}
    &f_i^{res} = \text{FC}(de_i'), \\
    &f_i = f_i^{dense}+f_i^{res}.
\end{align}

\emph{Designs of Scene Flow Predictor}: The scene flow predictor corrects the coarse scene flow by regressing residual flow from the refined flow embedding. In this paper, we adopt the concatenation of all five inputs of scene flow predictor and directly feed it into shared MLP to derive the refined flow embedding. On the other hand, FlowStep3D \cite{kittenplon2021flowstep3d} and PV-RAFT \cite{wei2021pv} propose to use a GRU-based gated activation unit on point clouds, inspired by RAFT \cite{teed2020raft}, for updating a hidden state. Given a hidden state $h_{l-1} \in \mathbb{R}^{c}$ from previous iteration and a current iteration vector $x_l$, $h_{l-1}$ is updated as follows:
\begin{align}
    \label{eq5}
    z_l &= \sigma(\text{SetConv}_{\text{z}}(h_{l-1} \oplus x_l)),\\
    r_l &= \sigma(\text{SetConv}_{\text{r}}(h_{l-1} \oplus x_l)),\\
    \widetilde{h_l} &=  \text{tanh}(\text{SetConv}_{\text{h}}((r_l \odot h_{l-1})\oplus x_l)),\\
    h_l &= (1-z_l)\odot h_{l-1}+z_l \odot \widetilde{h_l},
\end{align}
where $\sigma(\cdot)$ represents sigmoid activation function.

In particular, $x_l \in \mathbb{R}^{d_{re}+d_{pc_1}+d_{enc.}+3}$ is defined as the concatenation of flow re-embedding, the point feature of $PC_1$, the coarse dense flow, and the dense flow feature. We refer to the dense flow embedding as the hidden state that will be refined iteratively. Then, we can consider replacing our scene flow predictor with this newly designed GRU-based updating unit. Since FlowStep3D \cite{kittenplon2021flowstep3d} and PV-RAFT \cite{wei2021pv} claim that this GRU-based updating mechanism outperforms the fully-connected structure which is implemented in MLP, we will validate the performance of our scene flow predictor in our network architecture compared with this GRU-based method in the experiment.

\emph{Input of Scene Flow Predictor}: 
Another issue we want to investigate is what information is needed to predict the finer flow embedding. i.e. what information {is needed} in the input of the scene flow predictor. PointPWC-Net \cite{wu2020pointpwc} uses the flow re-embedding, the point feature of $PC_1$, the coarse dense flow, and the up-sampled coarse dense flow embedding as inputs. FlowStep3D \cite{kittenplon2021flowstep3d} additionally includes the dense flow feature but do not add the up-sampled flow embedding. PV-RAFT \cite{wei2021pv}, HCRF-Flow \cite{li2021hcrf}, and FESTA \cite{wang2021festa} all only use the flow re-embedding, the point feature of $PC_1$, and the coarse flow. HALFlow \cite{wang2021hierarchical} does not add the dense flow feature. In this sense, we explore whether removing certain elements from the input will degrade the performance and the extent to which those information contributes to the performance. We consider the flow re-embedding as an indispensable element because it encodes the motion between warped $PC_1$ and $PC_2$ in the current level.   

\emph{Level of Flow Refinement Layer}:
In \cite{wang2021hierarchical}, three flow refinement layers are applied for a more-for-less network architecture, which estimates $N$ points' scene flow from $4N$ points of raw input. {Only Three-NN interpolation is used for up-sampling in the finest level of flow estimation}. Apparently, this structure does not exhaustively leverage the functionality of the whole components in the flow refinement layer. Therefore, we choose to employ an additional flow refinement layer on the finest level to estimate the final scene flow. The above consideration raises another question: Can the performance be further improved if more
information from denser level is taken into account? Here we consider using raw point cloud with 4N points to estimate the 4N
points’ scene flow. The best choice will be demonstrated in the experiment.
\section{Experiments}
\subsection{Datasets and Data Preprocess}
Because of the inherent difficulty in acquiring large-scale ground-truth scene flow of the real world, we resort to the common synthetic FlyingThings3D dataset \cite{mayer2016large} for training and evaluation. We first train our network on FlyingThings3D dataset \cite{mayer2016large}, and then directly test our trained model on real-world LiDAR scans from KITTI scene flow dataset \cite{menze2018object} without any fine-tuning to demonstrate the generalization capability.

There are two common versions of preparing point clouds from FlyingThings3D dataset \cite{mayer2016large} and KITTI scene flow dataset \cite{menze2018object}. The first version of data preprocessing is proposed by HPLFlowNet \cite{gu2019hplflownet} and adopted in \cite{gu2019hplflownet,wu2020pointpwc,wang2021hierarchical,li2021hcrf,kittenplon2021flowstep3d}. It does not contain occlusion for input point clouds, which means each point in $PC_1$ has its corresponding point in $PC_2$ when warped by its ground-truth scene flow. The second version is proposed by FlowNet3D  \cite{liu2019flownet3d}. This version provides occluded point clouds as inputs and masks that indicate the invalid points without corresponding ones in the adjacent frame. These masks are also used in computing training loss and evaluation metrics. To compare with all 3D scene flow estimation methods to our knowledge, we follow two versions of data preprocessing and conduct experiments on both versions of datasets. More details about datasets and preprocessing is in the supplementary material.

\subsection{Training and Evaluation Details}
\subsubsection{Training Loss}
We train our network in a supervised manner at different levels, similar to \cite{sun2018pwc,wu2020pointpwc}. Suppose the predicted scene flow of each point at level $l$ is $\{f_i^l \in \mathbb{R}^3~|~ i=1,\dots,N_l\}$ and the ground-truth scene flow is $\{GT(f_i^l) \in \mathbb{R}^3~|~ i=1,\dots,N_l\}$. Here, $N_l$ denotes the number of points at level $l$. Our training loss can be therefore written as:
\begin{equation}
    Loss = \sum_{l=1}^4 \psi_l \frac{1}{N_l} \sum_{i=1}^{N_l} \left\|f_i^l-GT(f_i^l)\right\|_2,
\end{equation}
where $\psi_l$ indicates the weight at level $l$. We define the finest level, which is also the level with the densest points, as level $l=1$. Specifically, our network takes in $4N=8,912$ as inputs, $N_1=N=2,048$, $N_2=N / 2 =1,024$, $N_3 = N / 8 = 256$, and $N_4 = N / 32 = 64$. The loss weights are $\psi_1 = 0.02$, $\psi_2 = 0.04$, $\psi_3 = 0.08$, and $\psi_4 = 0.16$.
\subsubsection{Implementation Details}
For the training and evaluation process of our network, 8,192 points are randomly sampled as inputs from the raw points clouds of two consecutive frames. Only 3D XYZ coordinates of the point clouds are fed into our network, like \cite{gu2019hplflownet,wu2020pointpwc,wang2021hierarchical,li2021hcrf,kittenplon2021flowstep3d}. {For fair comparison with previous methods}, on FlyingThings3D prepared by \cite{gu2019hplflownet}, we first train our network on one quarter of the training set (4,910 pairs) and then fine-tune our model on the complete training set to speed up the training process. On FlyingThings3D prepared by \cite{liu2019flownet3d}, we train our model without fine-tuning.

We conduct all the experiments on a single Titan RTX GPU. Pre-training is done for 800 epochs, and fine-tuning lasts {for 200 epochs} after loading the pre-trained weights. Batch size is 14. Adam optimizer \cite{kingma2014adam} is used in training, and $\beta_1 = 0.9$, $\beta_2 = 0.99$. The initial learning rate is $0.001$ and decays for every $80$ epochs exponentially with decay rate $\gamma = 0.5$. Our supplementary material provides all details of network parameters. 
\subsubsection{Evaluation Metrics}
We adopt the same evaluation metrics used in \cite{gu2019hplflownet,wu2020pointpwc,wang2021hierarchical,li2021hcrf} to evaluate our model for fair comparison, including EPE3D(m), Acc3D Strict, Acc3D Relax, Outliers3D, EPE2D(px), and Acc2D. The detailed description of the metrics is shown in the supplementary material.
\section{Results}
\subsection{Comparison with State-of-the-Art (SOTA)}
Table \ref{table:flyingthing3d} shows the quantitative comparison between previous state-of-the-arts and our approach on FlyingThings3D dataset \cite{mayer2016large} and KITTI scene flow dataset \cite{menze2018object} prepared by \cite{gu2019hplflownet}. It is demonstrated that our approach outperforms all other methods by a large margin for both 3D and 2D metrics on FlyingThings3D dataset \cite{mayer2016large}. Meanwhile, our method also achieves the best generalization results on KITTI scene flow \cite{menze2018object}. Specifically, we surpasses the SOTA method, FlowStep3D \cite{kittenplon2021flowstep3d}, by 38.2\% for EPE3D metric on FlyingThings3D dataset \cite{mayer2016large}, and 43.4\% on KITTI scene flow \cite{menze2018object} dataset.

The recent work, FESTA \cite{wang2021festa}, is only tested on the datasets prepared by \cite{liu2019flownet3d}. To compare with all the methods to our knowledge, we also present the evaluation results on FlyingThings3D dataset \cite{mayer2016large} and KITTI scene flow dataset \cite{menze2018object} prepared by \cite{liu2019flownet3d} in Table \ref{table:flyingthing3d_by_flownet3d}. It can be demonstrated that our approach still outperforms previous methods substantially for all 3D metrics on both datasets.  
Specifically, we surpasses the SOTA method, FESTA \cite{wang2021festa}, by 43.2\% with respect to EPE3D metric on FlyingThings3D dataset \cite{mayer2016large}, and 24.7\% on KITTI scene flow \cite{menze2018object} dataset.
We believe the superior performance of our method on the datasets with occlusion partly from our backward validation, which can be aware of the occlusion in the network inference.   

\begin{figure*}[t]
	\centering
	\includegraphics[width=0.95\linewidth]{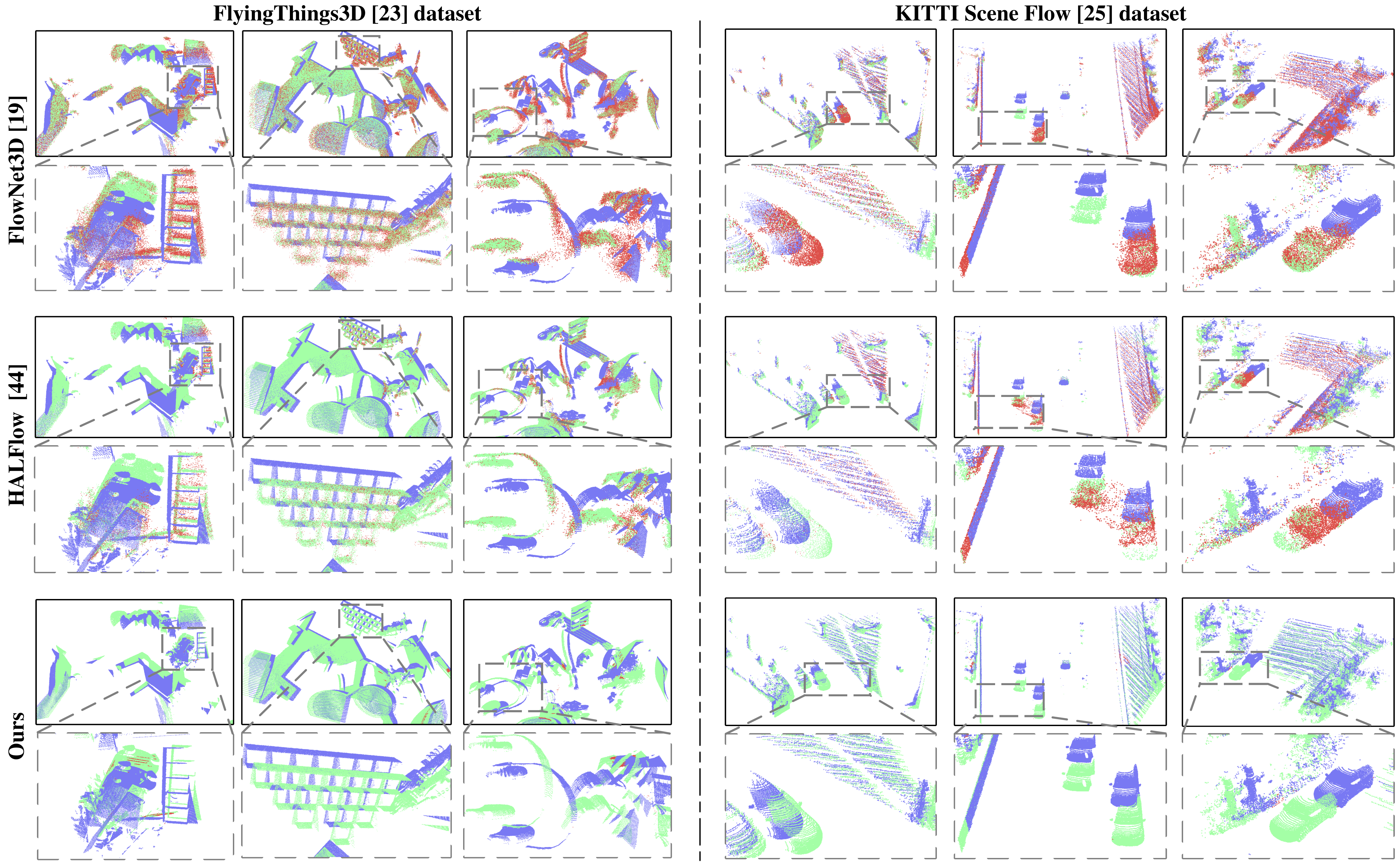}
	\caption{The visualization results of the proposed method, compared with FlowNet3D \cite{liu2019flownet3d} and HALFlow \cite{wang2021hierarchical}, on FlyingThings3D (left) and KITTI scene flow (right) datasets prepared by Gu \emph{et al}. \cite{gu2019hplflownet}. Blue points indicate $PC_1$. Green points indicate accurate predictions $\widetilde{PC_2} = PC_1 + F$ and red points indicate inaccurate predictions (measured by Acc3D Relax).  }   
	\label{figure:vis}
\end{figure*}

We also present detailed visualization of the accuracy of the estimated scene flow by our approach in Fig. \ref{figure:vis}, compared with methods in \cite{liu2019flownet3d,wang2021hierarchical}. It can be seen that our method can better handle the structures with repetitive patterns and large motions.  

\begin{table*}[!t]
	\begin{center}
		\caption{The quantitative comparison between recent state-of-the-art methods and ours on FlyingThings3D and KITTI scene flow datasets prepared by Gu \emph{et al}. \cite{gu2019hplflownet} without occlusion. All listed approaches are only trained on FlyingThings3D dataset. The best results are in bold. ``Full" means fully-supervised training.}
		\label{table:flyingthing3d}
		\resizebox{1.00\textwidth}{!}
            {
		\begin{tabular}{clccccccccc}
			\toprule
			Evaluation Dataset& Method&Training Data  &Input& Sup. & EPE3D & Acc3D Strict& Acc3D Relax & Outliers3D & EPE2D & Acc2D\\ \midrule
			& FlowNet3 \cite{ilg2018occlusions}   &  Quarter  & RGB stereo   & Full & 0.4570 & 0.4179& 0.6168& 0.6050 & 5.1348 &  0.8125\\ \cline{2-11}\noalign{\smallskip}
			
			& ICP \cite{besl1992method}   &  No    & Points & Full & 0.4062& 0.1614 & 0.3038  & 0.8796 & 23.2280 & 0.2913 \\
			& FlowNet3D \cite{liu2019flownet3d} &  Quarter  & Points & Full & 0.1136 & 0.4125 & 0.7706& 0.6016& 5.9740 & 0.5692 \\
			& SPLATFlowNet \cite{su2018splatnet}&  Quarter & Points & Full  & 0.1205 & 0.4197& 0.7180& 0.6187&  6.9759 & 0.5512 \\
			& HPLFlowNet \cite{gu2019hplflownet}   &   Quarter  & Points  & Full   & 0.0804 & 0.6144 & 0.8555& 0.4287& 4.6723 & 0.6764 \\
			& HPLFlowNet \cite{gu2019hplflownet}   &   Complete  & Points  &Full    & 0.0696 &  ---   & --- & ---& --- &   ---  \\
			& PointPWC-Net \cite{wu2020pointpwc}   &   Complete  & Points  & Full   & 0.0588   & 0.7379  & 0.9276 & 0.3424 & 3.2390 &  0.7994\\
			 &  HALFlow \cite{wang2021hierarchical}  &  Quarter & Points & Full  & 0.0511 &  0.7808 &  0.9437  & 0.3093  &  2.8739& 0.8056 \\ 
			&  HALFlow \cite{wang2021hierarchical}   &  Complete & Points  & Full &  0.0492  &  0.7850&  0.9468 & 0.3083 &2.7555 &  0.8111 \\
			&  FLOT \cite{puy2020flot}   &  Complete & Points & Full  & 0.0520  &  0.7320 & 0.9270 &  0.3570 & ---&  ---  \\
			\multirow{-9}{*}{\begin{tabular}[c]{@{}c@{}}FlyingThings 3D \\ dataset \cite{mayer2016large} \end{tabular}}
			
			&  HCRF-Flow \cite{li2021hcrf}      &  Quarter & Points & Full  & 0.0488 &0.8337  &  0.9507 &  0.2614  & 2.5652& 0.8704   \\
			
			& PV-RAFT  \cite{wei2021pv} & Complete& Points& Full & 0.0461 & 0.8169  & 0.9574  &  0.2924 & ---& ---    \\

			& FlowStep3D  \cite{kittenplon2021flowstep3d} & Complete& Points& Full & 0.0455 & 0.8162  & 0.9614  &  0.2165 & ---& ---    \\
			
			&  Ours    &  Quarter & Points & Full & 0.0317   &  0.9109&0.9757   & 0.1673 & 1.7436& 0.9108   \\ 
			
			&  Ours     &  Complete & Points & Full & \bf0.0281 & \bf0.9290 & \bf0.9817   &   \bf0.1458   &  \bf1.5229 &   \bf0.9279   \\ 
			
			\midrule
			& FlowNet3 \cite{ilg2018occlusions}  &  Quarter & RGB stereo  & Full & 0.9111 & 0.2039  & 0.3587 & 0.7463&  5.1023  & 0.7803 \\
			\cline{2-11}\noalign{\smallskip}
			
			& ICP \cite{besl1992method}   &  No    & Points &Full & 0.5181 & 0.0669 & 0.1667& 0.8712& 27.6752  & 0.1056 \\
			& FlowNet3D \cite{liu2019flownet3d}  &   Quarter & Points & Full  & 0.1767& 0.3738 & 0.6677  & 0.5271  &  7.2141  & 0.5093   \\
			& SPLATFlowNet \cite{su2018splatnet} &  Quarter & Points &Full & 0.1988  & 0.2174  & 0.5391 & 0.6575  &  8.2306  &  0.4189       \\
			& HPLFlowNet \cite{gu2019hplflownet}  &  Quarter  & Points & Full  & 0.1169 & 0.4783  & 0.7776  & 0.4103  &  4.8055  &  0.5938  \\
			& HPLFlowNet \cite{gu2019hplflownet}  &  Complete   & Points  & Full  & 0.1113 &   ---  &  ---  &  ---  &   ---  & --- \\
			& PointPWC-Net \cite{wu2020pointpwc}  &  Complete  & Points  &Full   & 0.0694 &  0.7281  &  0.8884 & 0.2648 &3.0062 &0.7673 \\
			&  HALFlow \cite{wang2021hierarchical}   &  Quarter & Points & Full&  0.0692&  0.7532&   0.8943 & 0.2529&  2.8660 &   0.7811 \\      
			&  HALFlow \cite{wang2021hierarchical}&  Complete   & Points  & Full &  0.0622 & 0.7649 &0.9026 & 0.2492 & 2.5140 &  0.8128 \\ 
			&  FLOT \cite{puy2020flot} &  Complete & Points & Full  & 0.0560 &  0.7550  &  0.9080  & 0.2420   & ---& --- \\
			\multirow{-9}{*}{\begin{tabular}[c]{@{}c@{}}KITTI \\ dataset \cite{menze2018object} \end{tabular}} 
			
			&  HCRF-Flow \cite{li2021hcrf}   &  Quarter & Points & Full  & 0.0531& 0.8631&   0.9444 &0.1797  &  2.0700& 0.8656 \\
			
			& PV-RAFT  \cite{wei2021pv} & Complete& Points& Full & 0.0560 & 0.8226  & 0.9372  &  0.2163 & ---& ---    \\
			
			&FlowStep3D \cite{kittenplon2021flowstep3d}  &  Complete &Points& Full& 0.0546 &0.8051  & 0.9254  & \bf0.1492  &  ---&     ---\\
			
			&  Ours &  Quarter & Points & Full & 0.0332  & 0.8931  & 0.9528  &  0.1690  & 1.2186& 0.9373 \\ 
			
			&  Ours   &  Complete & Points & Full & \bf0.0309 &  \bf0.9047  &\bf0.9580    &    0.1612   &    \bf1.1285 &       \bf0.9451  \\
			\bottomrule
		\end{tabular}
		}
	\end{center}
\end{table*}

\setlength{\tabcolsep}{4mm}
\begin{table*}[!t]
	\begin{center}
		\caption{Evaluation results on FlyingThings3D and KITTI Scene Flow datasets prepared by Liu \emph{et al}. \cite{liu2019flownet3d} with occlusion. The best results are in bold. ``Self" means self-supervised training. ``Full" means fully-supervised training. All fully-supervised approaches are trained on FlyingThings3D dataset. Self-Point-Flow \cite{li2021self} is trained on raw LiDAR data from KITTI dataset \cite{geiger2013vision}.}
		\label{table:flyingthing3d_by_flownet3d}
		\resizebox{1.00\textwidth}{!}
            {
		\begin{tabular}{clcccccc}
			\toprule
			Evaluation Dataset& Method  &Input & Sup. & EPE3D & Acc3D Strict& Acc3D Relax & Outliers\\ \midrule

			& FlowNet3D \cite{liu2019flownet3d} & Points &Full  & 0.169 & 0.254 & 0.579 & 0.789\\
			
			&  FLOT \cite{puy2020flot}    & Points &Full & 0.156  &  0.343 & 0.643  & 0.700 \\
			
			&  FESTA \cite{wang2021festa}    & Points &Full & 0.111  &  0.431 & 0.744  & --- \\
			
			\multirow{-4}{*}{\begin{tabular}[c]{@{}c@{}}FlyingThings 3D \\ dataset \cite{mayer2016large} \end{tabular}}
			
			&  Ours      & Points &Full & \bf0.063 & \bf0.791 & \bf0.909  & \bf0.279 \\ 
			\midrule
			&  Self-Point-Flow \cite{li2021self}  & Points &Self& 0.105 & 0.417&  0.725 &0.501 \\
			\cline{2-8}\noalign{\smallskip}
			& FlowNet3D \cite{liu2019flownet3d} & Points &Full & 0.173 & 0.276 & 0.609  & 0.649\\
			
			&  FLOT \cite{puy2020flot}    & Points &Full & 0.110  &  0.419 & 0.721   & 0.486\\
			
			&  FESTA \cite{wang2021festa}    & Points &Full & 0.097  &  0.449 & 0.833  & --- \\
			\multirow{-5}{*}{\begin{tabular}[c]{@{}c@{}}KITTI \\ dataset \cite{menze2018object} \end{tabular}}
			
			&  Ours      & Points&Full  &\bf0.073  &\bf0.819  & \bf0.890 & \bf0.261  \\ 
			\bottomrule
		\end{tabular}
		}
	\end{center}
\end{table*}

\subsection{Ablation Study} 
In this section, we demonstrate the effectiveness of the proposed all-to-all point mixture module and validate our network design choices compared with other discussed alternative structures through a series of ablation studies. All methods in the ablation studies are trained on $\frac{1}{4}$ of the training set (4,910 pairs) of FlyingThings3D dataset \cite{mayer2016large} prepared by \cite{gu2019hplflownet} and evaluated using the corresponding evaluation set. 
\subsubsection{All-to-all and Backward Information} In order to demonstrate the effectiveness of both the all-to-all mechanism and the backward validation vector, we first remove the vector that stores the backward information in the all-to-all flow embedding layer, which means $s^k_i$ is removed from formula (\ref{eq:similarity}). Then, the all-to-all mechanism is removed, which means points from $PC_1$ will only select $K$ nearest neighbours instead. The results in Table \ref{table:ablation}(a) show that both the all-to-all mechanism and the backward validation vector contribute to the improvement of performance. In fact, the all-to-all mechanism enables the querying point in $PC_1$ to expand its searching range from $K$ nearest neighbours to all points in $PC_2$ to determine its most reliable matching candidate. The backward validation vector acquired based on the all-to-all mechanism brings in the information that imposes bi-directional consensus from the backward direction. Therefore, the network can be guided by this backward validation information to better learn the correct matching and estimate more accurate scene flow.    

\subsubsection{Calculation of Point Similarity} Three different forms of similarity calculation are discussed in section \ref{section:all2all}. Since point similarity design can affect the correlation of adjacent point clouds to a large extent, we compare these product similarity methods with ours. The results in Table \ref{table:ablation}(b) demonstrate that the concatenation of point feature has the best performance than product forms of point similarity. Representing point similarity via concatenation is more suited in our network. We believe this is because the concatenation operation allows the network to fully exploit its self-learning ability. 

\subsubsection{Designs of Scene Flow Predictor}
Since our scene flow predictor is only implemented with MLP, we want to know whether GRU in \cite{kittenplon2021flowstep3d,wei2021pv} can improve the performance. We consider the flow embedding as the hidden state to be updated and replace our MLP based scene flow predictor with GRU. Table \ref{table:ablation}(c) show that the performance actually degrades using GRU. For our coarse-to-fine network, GRU is less suitable for updating and refining the flow embedding. We believe this is due to the difference of the point number at different levels, which is different from FlowStep3D \cite{kittenplon2021flowstep3d} and PV-RAFT \cite{wei2021pv}.   
\subsubsection{Input of Scene Flow Predictor} As the scene flow predictor serves to refine the flow embedding for regression of more accurate scene flow, we are interested in what matters in the input information of the scene flow predictor and to what extent each of the inputs contributes to the performance. Therefore, we respectively ablate each of the five inputs except the flow re-embedding in our scene flow predictor to demonstrate their importance. Table \ref{table:ablation}(d) show that the up-sampled flow embedding is the most important element of inputs, which is intuitive because the up-sampled flow embedding is refined in the previous layer and contains abundant motion information. The removal of the coarse flow and flow feature also causes a slight decline in quantitative performance, which demonstrates that they can also provide some instructions for the improvement of the refinement.   
\setlength{\tabcolsep}{0.9mm}
\begin{table*}[!t]
	 \caption{The ablation study results on FlyingThings3D dataset prepared by Gu \emph{et al}. \cite{gu2019hplflownet}.}
	\footnotesize
	\begin{center}
		\resizebox{1.0\textwidth}{!}
		{
	\begin{tabular}{l|l||cccc|cc}
				\toprule
				& Method&  EPE3D  &  Acc3D Strict   & Acc3D Relax & Outliers &  EPE2D & Acc2D  \\ 
				\noalign{\smallskip}
				\hline\hline
				\noalign{\smallskip}
				(a)   
				&Ours w/o backward validation    
				&0.0332 & 0.9044
				&0.9743 & 0.1766	
				& 1.8221 & 0.9065 
				
				\\
				& Ours w/o backward validation and all-to-all mechanism   
				& 0.0349& 0.9001
				&0.9725 & 0.1798
				& 1.9819 & 0.9032
				 
				\\
				
				&Ours (full, with backward validation and all-to-all mechanism)     
				&\bf0.0317 & \bf0.9109
				&\bf0.9757 & \bf0.1673
				&\bf1.7436 & \bf0.9108
				 
				\\
				\cline{1-8}\noalign{\smallskip}

				(b) &Ours (with product similarity)    
				&0.0356 &0.8872
				&0.9692 & 0.1953
				& 1.9872 &  0.8870
				
				\\
				
				&Ours (with cosine product similarity)      
				&0.0370 &0.8755
				&0.9670 & 0.2142
				& 2.0637 &  0.8746
				
				\\
				&Ours (with normalized product similarity )     
				&0.0339 & 0.8971
				&0.9724 & 0.1845
				&1.8790 & 0.8965
				
				\\
				&Ours (full, with concatenated similarity)     
				&\bf0.0317 & \bf0.9109
				&\bf0.9757 & \bf0.1673
				&\bf1.7436 & \bf0.9108
				
				\\
				\cline{1-8}\noalign{\smallskip}

				(c) 
				&Ours (replace Scene Flow Predictor with GRU)
				& 0.0350 &0.8892
				&0.9668 & 0.1827
				& 1.9274 & 0.8896 
				\\
				&Ours (full, with Scene Flow Predictor)     
				&\bf0.0317 & \bf0.9109
				&\bf0.9757 & \bf0.1673
				&\bf1.7436 & \bf0.9108
				\\
				
				\cline{1-8}\noalign{\smallskip}
				(d) 
					&Ours w/o features of $PC_1$ in Scene Flow Predictor  
				&0.0333 & 0.9047
				&0.9743 & 0.1740	
				& 1.8428 &  0.9073
				
				\\
				&Ours w/o up-sampled flow embedding in Scene Flow Predictor     
				&0.0380 & 0.8732 
				&0.9642 & 0.2099 
				&2.0953 & 0.8785
				
				\\
				&Ours w/o coarse flow in Scene Flow Predictor     
				&0.0323 & 0.9076
				&0.9750 & 0.1717 
				&1.7760 & 0.9083
				
				\\
				&Ours w/o flow feature in Scene Flow Predictor     
				&0.0327 & 0.9061
				&0.9748 & 0.1740
				&1.8063 & 0.9074
				
				\\
				
				&Ours (full, with complete five inputs in Scene Flow Predictor)     
				&\bf0.0317 & \bf0.9109
				&\bf0.9757 & \bf0.1673
				&\bf1.7436 & \bf0.9108
				      
				\\
	
				\cline{1-8}\noalign{\smallskip}
				(e) 
				&Ours (with interpolation estimating 2048 points' flow)
				& 0.0359 &0.8844 
				&0.9691 & 0.2004
				& 1.9511 & 0.8911 
				\\
				&Ours (with interpolation estimating 8192 points' flow)     
				&0.0332 & 0.9043
				&0.9739 & 0.1740
				&1.8039 & 0.9076
				\\
				&Ours (full, with flow refinement estimating 2048 points' flow)     
				&\bf0.0317 & \bf0.9109
				&\bf0.9757 & \bf0.1673
				&\bf1.7436 & \bf0.9108

				\\
				\bottomrule
			\end{tabular}
		}
	\end{center}
	\label{table:ablation}
\end{table*}

\subsubsection{Level of Flow Refinement Layer} In our hierarchical flow refinement module, four flow refinement layers exhaustively utilize all four levels of sampled point feature, indicated by skip connection. Compared with \cite{wang2021hierarchical}, we add an additional flow refinement layer on the finest level. In order to validate its effectiveness, we remove this flow refinement layer and instead use Three-NN interpolation to obtain the final $N$ points' scene flow from $4N$ points' inputs, as \cite{wang2021hierarchical}. In addition, we further apply the Three-NN interpolation to obtain $4N$ points' scene flow out of $4N$ points' inputs to see whether more information from raw input will result in an improvement. It turns out in Table \ref{table:ablation}(e) that both of these designs will degrade the network performance and our four layers of flow refinement that leverage all levels' sampled point feature prove the most suitable and effective structure design for our network. 
\section{Conclusion}
In this paper, a novel all-to-all point mixture module with backward reliability validation is proposed for reliable correlation. In addition, different designs and techniques for key components of our network are thoroughly compared. We provide a series of ablation studies to show the contributions of each element in key components and to demonstrate what matters in scene flow network. Quantitative results on FlyingThings3D \cite{mayer2016large} and KITTI scene flow dataset \cite{menze2018object} show that our method achieves SOTA performance. Our comparison and analysis on design choices of key components and structure are expected to facilitate the design of scene flow network in future research.

{\bf \small  Acknowledgement.} {\small This work was supported in part by the Natural Science Foundation of China under Grant 62073222, Grant U21A20480, and Grant U1913204; in part by the Science and Technology Commission of Shanghai Municipality under Grant 21511101900; and in part by the Open Research Projects of Zhejiang Laboratory under Grant 2022NB0AB01. The authors gratefully appreciate the contribution of Chaokang Jiang from China University of Mining and Technology, and Xinrui Wu from Shanghai Jiao Tong University. }
\clearpage
%
%
\bibliographystyle{splncs04}
\bibliography{egbib}
\end{document}